# Bridging Text and Vision: A Multi-View Text-Vision Registration Approach for Cross-Modal Place Recognition

Tianyi Shang[1,2], Zhenyu Li[1,*], Pengjie Xu[3], Jinwei Qiao[1], Gang Chen[4], Zihan Ruan[2], Weijun Hu[2]
https://github.com/nuozimiaowu/Text4VPR

***Abstract*— Mobile robots necessitate advanced natural language understanding capabilities to accurately identify locations and perform tasks such as package delivery. However, traditional visual place recognition (VPR) methods rely solely on single-view visual information and cannot interpret human language descriptions. To overcome this challenge, we bridge text and vision by proposing a multiview (360° views of the surroundings) text-vision registration approach called Text4VPR for place recognition task, which is the first method that exclusively utilizes textual descriptions to match a database of images. Text4VPR employs the frozen T5 language model to extract global textual embeddings. Additionally, it utilizes the Sinkhorn algorithm with temperature coefficient to assign local tokens to their respective clusters, thereby aggregating visual descriptors from images. During the training stage, Text4VPR emphasizes the alignment between individual text-image pairs for precise textual description. In the inference stage, Text4VPR uses the Cascaded Cross-Attention Cosine Alignment (CCCA) to address the internal mismatch between text and image groups. Subsequently, Text4VPR performs precisely place match based on the descriptions of text-image groups. On Street360Loc, the first text to image VPR dataset we created, Text4VPR builds a robust two baseline (two aligned methods: data alignment without without CCCA module and alignment using CCCA module), achieving a leading top-1 accuracy of 56% and a leading top-10 accuracy of 91% within a 5-meter radius on the test set, which indicates that localization from textual descriptions to images is not only feasible but also holds significant potential for further advancement, as shown in Figure 1.**

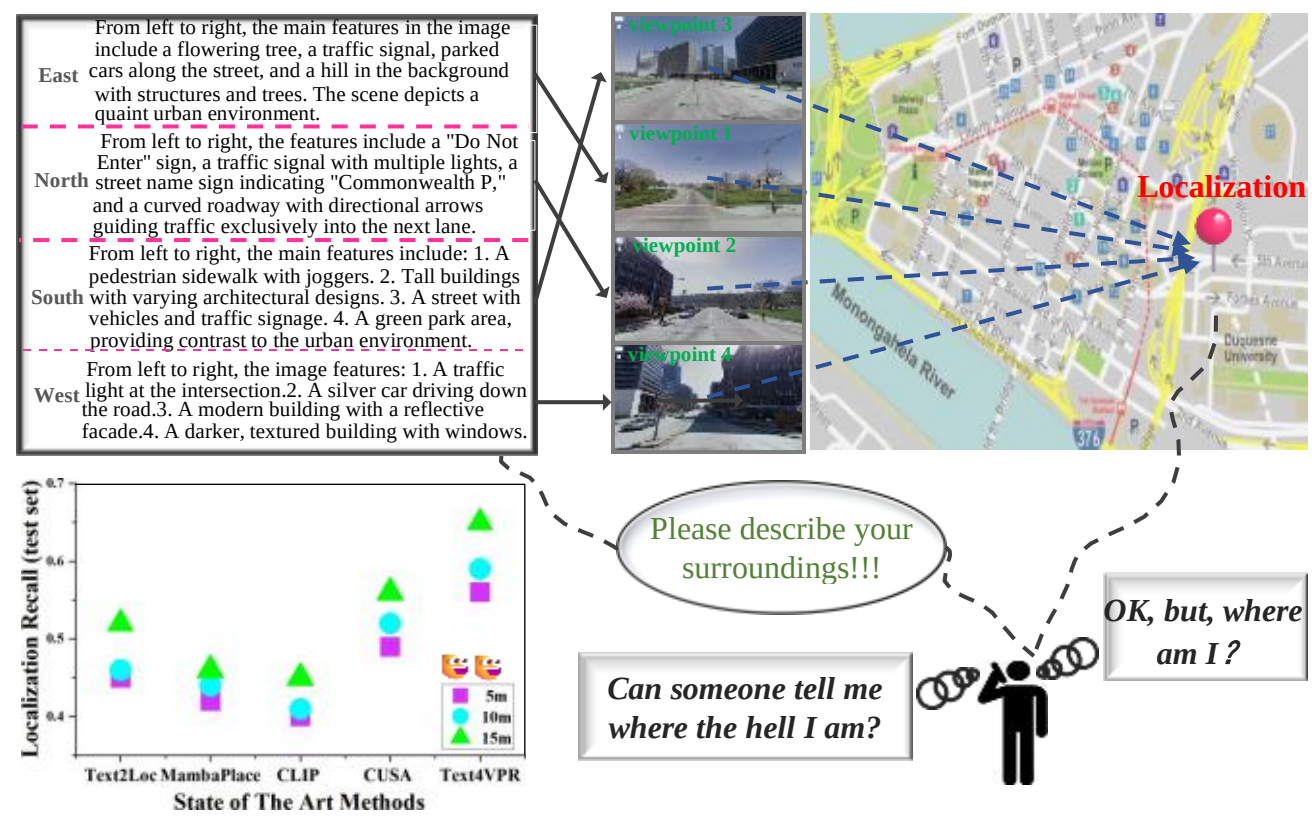

Fig. 1. We introduce Text4VPR, the first baseline for text-to-image place recognition task on self-built Street360Loc dataset. As a city-scale place localization solution,Text4VPR takes four text queries describing a location and retrieves four matching images from different views of the same place, ultimately identifying the most probable location. The proposed Text4VPR implementation consistently achieves better performance on all top semantic-based localization methods.

## I. Introduction

Traditional self-localization systems typically support unimodal queries (text, image, or point cloud) [1], [2]. Advanced systems should enable cross-modal queries, allowing users to express concepts that guide the robot to localization more effectively.

Future intelligent robots should collaborate with humans to utilize large-scale scenes and natural language descriptions for self-localization in tasks like goods delivery and vehicle pickup. This cross-modal approaches improve the ability of robots to navigate in complex urban environments. For example, a delivery robot with natural language processing can use a human's verbal description for scenes to find the destination. However, challenges include the high computational complexity of processing large point cloud data in text-to-point cloud cross-modal schemes, which can slow down operations. Additionally, in text-to-image cross-modal schemes, the lack of multi-view scene descriptions limits contextual cues for self-localization.

To address these challenges, we propose a multi-view text-to-image place recognition approach called Text4VPR. During the training stage, we focus on aligning individual images with their corresponding textual descriptions. Specifically, the text is first input into a frozen T5 model, followed by an attention layer that generates compact text vectors. The images are initially encoded by the ViT [3] to extract local tokens, and the Sinkhorn algorithm with a learnable temperature coefficient is applied to cluster these tokens, ultimately producing compact image vectors. The temperature coefficient can be viewed as a filter that selectively disregards tokens that are either unimportant or lack significant discriminative power for the image. Finally,

*This work was supported by the Natural Science Foundation of Shandong Province (ZR2024QF284), the Opening Foundation of Key Laboratory of Intelligent Robot (HBIR202301), the Open Project of Fujian Key Laboratory of Spatial Information Perception and Intelligent Processing (FKLSIPIP1027). *(Corresponding authors: Zhenyu Li)*

[1]Tianyi Shang, Zhenyu Li, and Jinwei Qiaoare with the School of Mechanical Engineering, Qilu University of Technology (Shandong Academy of Sciences), Jinan 250353, China (832201319@fzu.edu.cn; lizhenyu@qlu.edu.cn)

[2]Tianyi Shang, Zihan Ruan, and Weijun Hu are with the Department of Electronic and Information Engineering, Fuzhou University, Fuzhou 350100, China (832201319@fzu.edu.cn; ZIHAN.RUAN.2023@mumail.com; 949920389@qq.com)

[3]Pengjie Xu is with the School of Mechanical Engineering, Shanghai Jiao Tong University, Shanghai 200030, China (xupengjie194105@sjtu.edu.cn)

[4]Gang Chen is with the Xiamen University, Xiamen, China (gangchen9704@163.com)

these extracted image and text features are optimized by contrastive learning.

In the inference stage, our focus shifts to aligning groups of descriptions with corresponding groups of images. Considering that the order of description groups and image groups for the same location may not be aligned in real-world scenarios, we propose the CCCA method: after employing cascaded cross-attention to fuse text and image embeddings, we consider both the distance relationships in the feature space and the deep dynamic semantic relations to compute the image-text similarity, ultimately achieving precise alignment.

We present the first text-to-image localization dataset, named Street360Loc. Street360Loc is derived from the Google Street View dataset[4], in which 360° spherical imagery is segmented into four side views and one upward view. Considering that the upward view contains limited information, we only use the four side views to construct our dataset. By leveraging the emerging cross-modal large language model, ChatGPT [5], we generated descriptions for the four side views of each location, building the Street360Loc dataset. We then proofread and revised each description to ensure their correctness and alignment with natural human description patterns.

In summary, our primary contributions are as follows:

- We take the initial step in addressing the text-to-image cross-modal place recognition problem by introducing a robust single-stage baseline, Text4VPR, and present Street360Loc[1], the first dataset specifically designed for this task with dense semantic information.
- Using innovative CCCA module for text-vision registration, Text4VPR improves deep semantic relationships while preserving spatial distance relationships, effectively addressing internal misalignment between descriptions and image groups.
- Text4VPR employs the Sinkhorn algorithm in conjunction with a learnable temperature parameter to assign local image tokens to clusters, producing comprehensive image representations.
- Text4VPR employs individual text-image pairs for training and matches sets of texts with groups of images during inference. The training strategy enables the model to effectively learn both text and image features, while the inference strategy leverages multiview information for precise localization.

## II. Related Work

### A. Visual Place Recognition

The primary objective of VPR is to compare the image of the current location with an existing environmental image database to determine the position of an object accurately. Typically, these methods utilize pre-trained visual backbone networks, such as ResNet [6], ViT [3] and Swin Transformer [7] to extract visual descriptors from images. Subsequently, various aggregation techniques are employed to generate a global vector from these image descriptors. This highly aggregated global vector enhances the matching process between query images and database images. NetVLAD [8] was a pioneering method that introduced the VLAD [9] technique for aggregating image features. MixVPR [10] employs multilayer perceptrons (MLPs) to extract descriptions from various feature maps within the ResNet architecture. Recently, Sergio Izquierdo and colleagues proposed SALAD [11], which takes into account the dual matching relationship between clusters and feature information, introducing a ”trash bin” mechanism to create a more comprehensive global descriptor. However, global descriptors inevitably result in a loss of significant image detail. Consequently, many methods adopt a two-stage strategy: first, global descriptors are utilized to filter out the top-k candidates; then, local descriptors are employed to re-rank these top-k results. For instance, R2Former [12] leverages token information between image pairs for reranking.

Recently, VPR problems that utilize semantic information for localization have become a prominent area of research. Almost all methods employ sparse semantic descriptions to search through 3D point cloud map databases. Text2Pos [13] was a pioneering work in this field, which segments point clouds into different sub-maps and performs coarse-to-fine retrieval. Building on Text2Pos, RET [14] introduced transformers to enhance the representations of text and point clouds. Text2Loc [15] incorporates contrastive learning in the coarse stage and employs a matching-free method in the fine localization stage, significantly reducing computational overhead. The recent MambaPlace [16] utilized a selective State Space Model (SSM) mechanism to comprehensively represent large-scale point clouds, leading to improved accuracy.

However, the language descriptions utilized in the aforementioned methods are based on 21 static object labels, offering simplistic descriptions of up to three objects per location. Such sparse language descriptions are inadequate for addressing localization challenges in complex environments. In contrast, our method employs comprehensive language descriptions and is the first to tackle the text-to-image VPR problem.

### B. Vision-Language Cross-Modal Matching

The primary objective of text-image cross-modal matching is to assess the similarity between text and images, which significantly influences research areas such as cross-modal retrieval [17], image captioning[18], and text-to-image synthesis [19]. The most relevant area to our work is image-text cross-modal retrieval [20]. The standard approach to image-text cross-modal retrieval utilizes deep neural networks to hierarchically extract features from both images and texts, ultimately producing compact representations for each modality. Various methods are subsequently employed to assess their similarity and alignment, as demonstrated by Faghri et al. [21].

The above methods focus on aligning the overall text with the overall image. Lee et al. [22] tilized stacked attention mechanisms to explore the correspondence between salient objects in descriptions and images, achieving state-of-the-art results while enhancing interpretability. The SGRAF network [23] considered both global and local features, using graph convolutional neural networks to infer image-text similarity. CUSA [24] leverages the power of uni-modal pre-trained models to provide soft-label supervision signals. CLIP [25] demonstrates remarkable performance by using contrastive learning on large-scale datasets.

[1]https://github.com/nuozimiaowu/Text4VPR.

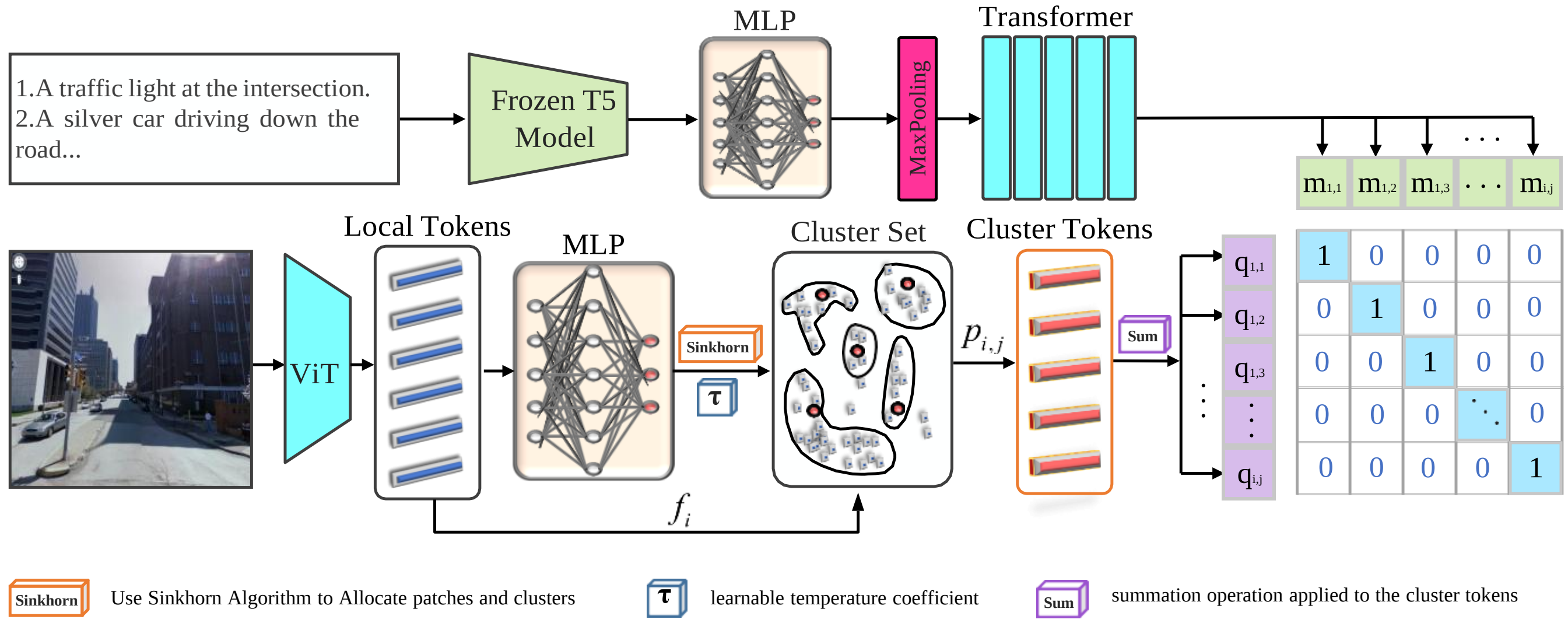

Fig. 2. In the training stage, Text4VPR utilize the T5 model to encode text descriptions and use the ViT encoder for visual encoding. Then we implement the *Sinkhorn* algorithm to assign tokens to clusters in the image, followed by cluster aggregation to derive the image encoding. Finally, we employ contrastive learning to bring correctly matched image-text pairs closer together.

These image-text studies demonstrate a strong semantic correlation between images and text, suggesting that they can be effectively semantically aligned. Inspired by this, our work is the first to address the text-to-image place recognition problem, focusing on retrieving images from text for accurate localization.

## III. Methodology

In this section, we present a multi-view approach for cross-modal place recognition, termed Text4VPR. Text4VPR employs contrastive learning to minimize the distance between encoded individual image-text pairs during the training stage. During the inference stage, it conducts multi-view matching, ultimately achieving precise localization based solely on text-to-image correspondence.

### A. Text Encoder

Each location in our dataset is represented by four different textual descriptions, corresponding to four different viewpoints. To encode these descriptions into meaningful feature representations, we define $D_i = \{d_{i,1}, d_{i,2}, d_{i,3}, d_{i,4}\}$ as the set of textual descriptions for location $i$, where each element represents a description of location $i$ facing a certain direction. We first process each description $d_{i,j}$ through a pre-trained text encoder parameterized by $\theta_t$, obtaining a feature vector:

$$t_{i,j} = f(d_{i,j}; \theta_t) \tag{1}$$

Here, we use the pre-trained T5 model as the encoder, and the parameters $\theta_t$ of the T5 model are entirely frozen. This transformation yields a fixed-length embedding $t_{i,j}$ that contains rich semantic information.

Subsequently, we employ an attention mechanism to capture the high-level features of the textual descriptions. We input the feature vector $t_{i,j}$ into a MaxPooling layer and an MLP to obtain a fixed-length vector:

$$a_{i,j} = MLP(MaxPool(t_{i,j})) \tag{2}$$

Next, we feed $a_{i,j}$ into a Transformer block, where the new $m_{i,j} = Transformer(a_{i,j})$ captures long-distance relationships between sentences, ultimately obtaining the global textual features. This results in $M_i = \{m_{i,1}, m_{i,2}, m_{i,3}, m_{i,4}\}$, representing the final encoded set of textual descriptions for location $i$.

### B. Image Encoder

As shown in Fig.2, we use a frozen ViT encoder for image encoding, transforming the input image $x$ into a set of tokens, represented as:

$$T = \{T_{global}, T_1, T_2, \ldots, T_n\} \tag{3}$$

Where $T_{global}$ is the global token, and $T_i$ represents the $i^{th}$ local token. We aim to assign the local tokens $T_n$ to image clusters to obtain a compact image representation. To achieve this, we merge the last two dimensions of $T_n$ and use a fully connected layer to reduce the dimensionality of $T_n$:

$$s_i = W_{s2}(\sigma(W_{s1}(t_i) + b_{s1})) + b_{s2} \tag{4}$$

This dimensionality reduction transforms $T_n$ into a shape of $[token_{num}, clusters]$, implicitly constructing a cost matrix $S$ between tokens and clusters. The cost matrix $S$ represents the distance from each local token to each cluster. To find the optimal assignment of local tokens to clusters, we model this allocation problem as an optimal transport problem and then solve it using the *Sinkhorn* algorithm. Then, we initialize two vectors $u$ and $v$, and update them iteratively during the process. The update rules for each iteration can be represented as:

$$u = \log a - \log \quad \exp(S_{reg} + v) \tag{5}$$

$$v = \log b - \log \quad \exp(S_{reg} + u) \tag{6}$$

Where $\log a$ and $\log b$ represent the normalized source and target weights, respectively. The regularized cost matrix $S_{reg}$ is defined as $S/reg$, where $reg$ is the regularization parameter." After multiple iterations, we can obtain the transport matrix $P$.

$$P = exp(\frac{S_{reg} + u + v}{\tau}) + exp(S_{reg} + u + v) \tag{7}$$

The transport matrix $P$ is the assignment matrix from $T_n$ to clusters. We introduce a learnable temperature parameter $\tau$ to adjust the 'softness' of the assignment, thereby alleviating issues where some local tokens cannot be effectively matched to clusters. We compute the weighted sum of all local tokens to obtain the final representation of each cluster. Defining the feature representation of each local token as $f_i$, and the aggregated feature $F_j$ of each cluster is represented as:

$$F_j = \quad p_{i,j} \cdot f_i \tag{8}$$

Finally, we perform flattening and normalization on the features $F_j$ to obtain the global descriptor vector of the entire image. We then define $Q_i = \{q_{i,1}, q_{i,2}, q_{i,3}, q_{i,4}\}$ as the set of multi-view images after encoding, where each $q_{i,j}$ is the global descriptor generated through the above process.

### C. Training Strategy

During the training stage, we do not consider the relationship between the text and image sets for a single location. Instead, we focus on the correlation between individual text-image pairs. As illustrated in Fig. 2, we utilize a contrastive learning-based loss function designed to enhance the similarity of correctly matched text-image pairs while diminishing the similarity of mismatched negative samples.

For any positive text-image pair $(m_{i,j}, q_{i,j})$, the loss function is represented as:

$$L_{train} = -\log \frac{\exp(sim(m_{i,j}, q_{i,j})/r)}{\sum_{k=1}^{N} \exp(sim(m_{i,j}, q_{i,j})/\tau)} \tag{9}$$

Where $sim(m_{i,j}, q_{i,j})$ denotes the similarity between the text descriptor $m_{i,j}$ and the image descriptor $q_{i,j}$, $\tau$ is the temperature parameter, $N$ is the number of samples in the batch.

### D. Inference Strategy

During the inference stage, we treat features from various viewpoints of the same location as a unified representation for location retrieval. Since the text-image datasets for each location are inherently misaligned, we first apply Cascaded Cross-Attention Cosine Alignment (CCCA) to achieve alignment, as shown in Fig. 3.

Given two inputs, $M_i$ and $Q_i$, we first generate an internally permuted set, denoted as $Q_{permuted}$, which contains the correctly ordered subset $Q_{aligned}$. Subsequently, the similarity between $M_i$ and every permuted element $Q_p \in Q_{permuted}$ is calculated. We first fuse the features of $M_i$ and $Q_p$ through a cascaded cross-attention mechanism. Specifically, $M_i$ and $Q_p$ alternately serve as the *query* and *key* in two consecutive attention layers, formulated as:

$$H = \text{CrossAtten}_N(Q_p, CrossAtten(M_i, Q_p)) \tag{10}$$

The generated $H$ represents dynamic interaction features, which encompass the co-occurrence information between $M_i$ and $Q_p$, capturing the fine-grained associations between text and images. Then, the similarity between $M_i$ and $Q_p$ can be defined as:

$$S_{sim} = Cos(Q_p, H) + Cos(Q_p, M_i) + Cos(M_i, H) \tag{11}$$

$Cos(Q_p, M_i)$ represent the spiral distance between $Q_p$ and $M_i$ in the feature space, while $Cos(Q_p, H)$ catch the dynamic fine-grained similarity. The combination of these two $Cos$ not only considers the distance relationship of the overall feature vectors in the space but also reflects the discriminative local dynamic interaction features, providing a more comprehensive representation of the *Similarity*. We compute the element with highest *Similarity* from $Q_{permuted}$ and treat it as $Q_{aligned}$, thus achieving the alignment.

Since the text-image datasets for each location have been aligned after CCCA, we treat features from various viewpoints of the same location as a unified representation for location retrieval. Specific, we directly concatenate the descriptors of the text and images from the same location, which is shown as:

$$M'_i = m_{i,1} \oplus m_{i,2} \oplus m_{i,3} \oplus m_{i,4} \tag{12}$$

$$Q'_i = q_{i,1} \oplus q_{i,2} \oplus q_{i,3} \oplus q_{i,4} \tag{13}$$

Where $\oplus$ denotes the concatenation operation of vectors. We then utilize these high-dimensional vectors for matching retrieval, enhancing accuracy and robustness. This method effectively integrates multi-view information from the same location, ensuring the model maintains high performance when confronted with a restricted field of view.

## IV. Experiment

### A. Experiment Setup

We train and evaluate Text4VPR using the custom-built Street360Loc dataset comprising 7,000 locations. Each location includes four images captured from different orientations, accompanied by corresponding textual descriptions. The distance between two adjacent locations in the Street360Loc dataset is approximately 6 meters. We partitioned the dataset into 5,000 locations for training, 1,000 for testing, and 1,000 for validation. To comprehensively assess Text4VPR's performance on the Street360Loc dataset, we employed top-k retrieval accuracy ($k = 1, 5, 10$) as an evaluation metric, indicating whether the correct location is included among the top-k retrieved results. Additionally, we introduce various error thresholds ($\varepsilon = 5m,\ 10m,\ 15m$), considering a retrieved location to be a correct match if it falls within these distance thresholds. This combination of $k$ and $\varepsilon$ values enables a thorough evaluation of Text4VPR's effectiveness under different conditions. We begin by pre-initialising the ViT encoder on a Tesla V100 GPU, configuring it to run for 10 epochs with a batch size of 64 and a learning rate of $1e^{-4}$, utilizing the Adam optimizer to ensure stable convergence. It is imperative to elucidate that we have delineated two methodologies for place recognition modal registration. The first approach employs a data alignment strategy, eschewing the utilization of the CCCA module, to effectuate an approximate registration of images and corresponding textual descriptors within the dataset. The second methodology, in contradistinction, orchestrates a cascaded feature-level registration of the visual and textual elements, meticulously facilitated by the CCCA module, thereby ensuring a more refined and precise alignment.

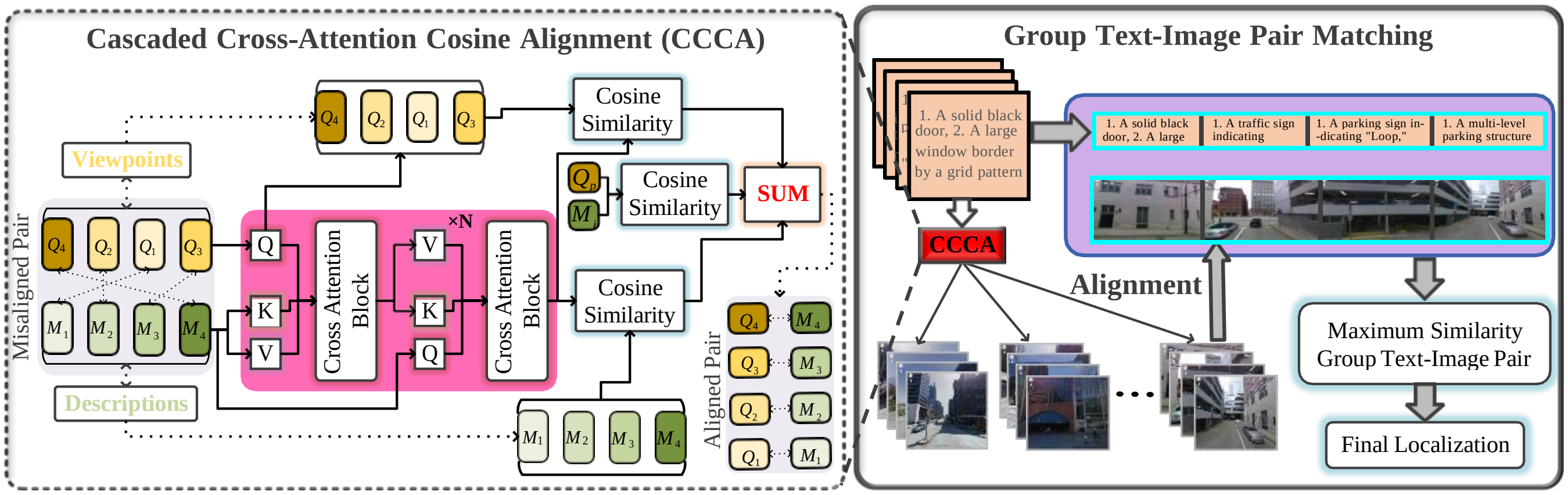

Fig. 3. In the inference stage, we first employ Cascaded Cross-Attention Cosine Alignment (CCCA) to internally align multi-view text description sets with image sets. Subsequently, we concatenate the four aligned text descriptions and their paired multi-view images into a unified descriptor, which is then used for text-image retrieval to achieve final robust localization.

### B. Performance Evaluation by Comparison with SOTA Tasks

We evaluate the performance of Text4VPR by benchmarking it against state-of-the-art (SOTA) methods on related tasks, as presented in Table I. To facilitate a fair comparison with text to point cloud place recognition methods in the Street360Loc dataset, we substituted PointNet with a pre-trained ViT for image feature extraction, while preserving the coarse-to-fine localization strategy employed in these methods. Another task related to ours is text-image retrieval, which can be directly tested in Street360Loc. To ensure a fair comparison, we used the same CCCA registration module and multi-view matching retrieval strategy across all methods compared. The results unequivocally confirm that our method surpasses previous approaches on the text based place recognition tasks, underscoring its superior performance.

### C. Performance Evaluation with Different Training Strategies

We further evaluate the impact of different training strategies on the performance of text-image place recognition. As shown in Table II, training with individual image-text pairs yields significantly higher accuracy compared to a group-based image-text training strategy. This improvement likely stems from the ability of individual image-text pairs to allow the image and text encoders to more effectively capture the unique features of each image and its corresponding descriptions. Consequently, these findings provide strong empirical support for the effectiveness of distinct training and inference strategies in Text4VPR.

### D. Ablation study

*1) Ablation Study with Text Encoder:* In the text encoding pipeline, textual descriptions are initially processed by a frozen T5 model to generate embedding vectors. Following this, we conduct an ablation study to evaluate subsequent processing methods, as detailed in Table III. The ”MLP + MaxPooling” operation projects these embeddings into a higher-dimensional space, while a Transformer layer captures essential features and long-range dependencies within the text.

Experimental results from Table III indicate that including the T1 module leads to a performance decline, likely because the Transformer disrupts the detailed information extracted by the T5 model. Based on these findings, we adopt a method that first applies ”MLP + MaxPooling” directly to the embeddings, followed by a Transformer layer to capture long-range positional relationships across sentences, producing a more comprehensive encoding of textual features.

*2) Ablation Study with Image Encoder:* As shown in Table IV, we demonstrate that image encoding based on optimal transport-based cluster assignment effectively improves model performance, followed by an evaluation of the role of the temperature coefficient. The decrease in accuracy after removing the temperature coefficient occurs because the

TABLE I

PERFORMANCE EVALUATION BY COMPARISON WITH SOTA PLACE RECOGNITION(WITH *) AND TEXT-IMAGE RETRIEVAL METHODS.

| Methods | Localization Recall ($\varepsilon < 5/10/15m$)↑ | | | | | |
|---|---|---|---|---|---|---|
| | Validation Set | | | Test Set | | |
| | k = 1 | k = 5 | k = 10 | k = 1 | k = 5 | k = 10 |
| Text2Pos*[13] | 0.28/0.29/0.34 | 0.29/0.33/0.34 | 0.37/0.39/0.41 | 0.21/0.23/0.29 | 0.27/0.29/0.30 | 0.35/0.37/0.38 |
| Text2Loc*[15] | 0.48/0.50/0.55 | 0.63/0.65/0.67 | 0.83/0.84/0.86 | 0.45/0.46/0.52 | 0.60/0.62/0.63 | 0.82/0.83/0.85 |
| MambaPlace*[16] | 0.50/0.50/0.56 | 0.62/0.63/0.65 | 0.86/0.87/0.88 | 0.42/0.44/0.46 | 0.56/0.58/0.58 | 0.81/0.82/0.83 |
| CLIP (fine-tuned) [25] | 0.44/0.45/0.48 | 0.58/0.59/0.60 | 0.85/0.87/0.87 | 0.40/0.41/0.45 | 0.54/0.55/0.56 | 0.80/0.81/0.82 |
| CUSA [24] | 0.58/0.60/0.64 | 0.83/0.85/0.85 | 0.90/0.90/0.92 | 0.49/0.52/0.56 | 0.80/0.82/0.83 | 0.87/0.88/0.90 |
| Text4VPR (CCCA alignment) | **0.64/0.66/0.72** | **0.88/0.88/0.90** | **0.94/0.95/0.96** | **0.56/0.59/0.65** | **0.85/0.86/0.89** | **0.91/0.92/0.94** |
| Text4VPR(data alignment) (our) | **0.65/0.67/0.74** | **0.89/0.88/0.91** | **0.95/0.96/0.96** | **0.57/0.60/0.66** | **0.86/0.87/0.89** | **0.92/0.93/0.94** |

TABLE II

PERFORMANCE EVALUATION ON TEST SET WITH DIFFERENT TRAINING STRATEGIES. "GROUP" DENOTES THE STRATEGY WHERE WE CONCATENATE MULTIPLE TEXTUAL DESCRIPTIONS AND IMAGE DESCRIPTORS FROM THE SAME LOCATION AND PERFORM CONTRASTIVE LEARNING, WHEREAS "SINGLE" REPRESENTS THE APPROACH WHERE INDIVIDUAL IMAGE-TEXT PAIRS ARE USED FOR CONTRASTIVE LEARNING. THE INFERENCE STRATEGY REMAINS CONSISTENT ACROSS BOTH TRAINING STRATEGIES.

| Training Strategy | Localization Recall ($\varepsilon < 5/10/15m$)↑ | | |
|---|---|---|---|
| | k = 1 | k = 5 | k = 10 |
| Group | 0.24/0.26/0.31 | 0.53/0.54/0.62 | 0.65/0.67/0.74 |
| Single | **0.56/0.59/0.65** | **0.85/0.86/0.89** | **0.91/0.92/0.94** |

TABLE III

ABLATION STUDIES FOR KEY COMPONENTS ON THE TEXT ENCODING STAGE. HERE, "M" DENOTES THE "MLP + MAXPOOLING" OPERATION, "T1" REPRESENTS ADDING A TRANSFORMER LAYER BEFORE THE "MLP + MAXPOOLING" OPERATION, AND "T2" INDICATES ADDING A TRANSFORMER LAYER AFTER THE "MLP + MAXPOOLING" OPERATION.

| Methods | Localization Recall ($\varepsilon < 5/10/15m$)↑ | | |
|---|---|---|---|
| | k = 1 | k = 5 | k = 10 |
| T1+M+T2 | 0.50/0.60/0.63 | 0.81/0.85/0.87 | 0.87/0.91/0.92 |
| T1+M | 0.41/0.45/0.50 | 0.72/0.75/0.79 | 0.81/0.83/0.87 |
| M | 0.44/0.47/0.53 | 0.74/0.77/0.82 | 0.84/0.86/0.89 |
| Text4VPR(M+T2) (CCCA alignment) | **0.56/0.59/0.65** | **0.85/0.86/0.89** | **0.91/0.92/0.94** |
| Text4VPR(M+T2) (data alignment) | **0.57/0.60/0.66** | **0.86/0.87/0.89** | **0.92/0.93/0.94** |

TABLE IV

ABLATION STUDIES FOR SINKHORN ALGORITHM AND KEY VARIABLE (TEMPERATURE COEFFICIENT $\tau$) ON THE IMAGE ENCODING PROCESSING. VIT + MAXPOOLING MEANS WE USE MAXPOOLING TO AGGRESSIVE TOKENS FROM VIT BACKBONE.

| Methods | Localization Recall ($\varepsilon < 5/10/15m$)↑ | | |
|---|---|---|---|
| | k = 1 | k = 5 | k = 10 |
| ViT+Maxpooling | 0.38/0.40/0.40 | 0.58/0.59/0.60 | 0.77/0.77/0.79 |
| w/o $\tau$ | 0.50/0.54/0.60 | 0.80/0.82/0.84 | 0.89/0.90/0.92 |
| Text4VPR (CCCA alignment) | **0.56/0.59/0.65** | **0.85/0.86/0.89** | **0.91/0.92/0.94** |
| Text4VPR(M+T2) (data alignment) | **0.57/0.60/0.66** | **0.86/0.87/0.89** | **0.92/0.93/0.94** |

TABLE V

ABLATION STUDIES FOR CCCA MODEL. W/O CCCA MEANS THAT THE CCCA MODULE IS NOT USED DURING INFERENCE. W/O CASCADE MEANS THAT ONLY THE COSINE SIMILARITY BETWEEN THE ENCODED TEXT AND IMAGES IS USED IN THE CCCA. W/O COSINE MEANS THAT ONLY THE SIMILARITY BETWEEN THE ENCODED TEXT AND THE TEXT-IMAGE MIXED REPRESENTATION IS USED IN THE CCCA. *Text4VPR(Aligned)* REFERS TO THE AUTOMATIC ALIGNMENT OF IMAGE AND TEXT GROUPS WITHIN THE DATABASE TO SIMULATE THE PERFORMANCE WHEN THE ALIGNMENT ACCURACY IS 100%, THEREBY EVALUATING THE ALIGNMENT CAPABILITY OF THE CCCA.

| Methods | Localization Recall ($\varepsilon < 5/10/15m$)↑ | | |
|---|---|---|---|
| | k = 1 | k = 5 | k = 10 |
| w/o CCCA | 0.15/0.15/0.16 | 0.20/0.22/0.23 | 0.30/0.32/0.33 |
| w/o Cascade | 0.52/0.56/0.63 | 0.82/0.83/0.83 | 0.89/0.91/0.91 |
| w/o Cosine | 0.53/0.54/0.62 | 0.82/0.82/0.84 | 0.90/0.90/0.91 |
| Text4VPR (CCCA alignment) | **0.56/0.59/0.65** | **0.85/0.86/0.89** | **0.91/0.92/0.94** |
| Text4VPR(data alignment) | **0.57/0.60/0.66** | **0.86/0.87/0.89** | **0.92/0.93/0.94** |

temperature coefficient enables smoother assignment of image tokens to clusters, mitigating the impact of mismatches where certain tokens may not align closely with any cluster.

*3) Ablation Study for CCCA:* To validate the effectiveness of CCCA in multi-view text-image alignment, we conduct extensive ablation experiments. As shown in Table V, in the absence of CCCA alignment, viewpoint confusion severely degrades the model's performance. We then consider the ideal case where images and texts are automatically aligned within the database. The results demonstrate that the performance after applying CCCA for viewpoint matching closely resembles this ideal case, strongly validating the effectiveness of our CCCA approach.

We further conduct ablation studies within CCCA, where using only the cosine similarity between text and image representations, or between text and mixed feature representations, resulted in a drop in model accuracy. This is because cosine similarity between text and image representations measures the spatial distance between the two encodings, whereas cosine similarity between text and mixed features implicitly captures the fine-grained similarity of dynamic, jointly attended parts between text and image. These two components complement each other, and both are essential for achieving accurate multi-view text-image alignment.

## *E. Robustness Study*

*1) Robustness Study with Different Description Preserve Ways:* In the previous section, we validate the effectiveness of each module through ablation studies. In this section, we investigate the model's robustness by progressively truncating input text descriptions to observe the effect on performance.

We start by reducing each textual description to varying lengths, using the truncated text for training and inference. As shown in Table VI, when only 25% of text description is

TABLE VI

ROBUSTNESS STUDY WITH DIFFERENT LENGTH OF EACH DESCRIPTION BASED ON THE TEST SET. WE PRESENT THE MODEL ACCURACY AFTER SEGMENTING EACH TEXTUAL DESCRIPTION. THE "25%" DESIGNATION INDICATES THAT ONLY THE FIRST 25% OF EACH TEXTUAL DESCRIPTION IS RETAINED, AND SIMILARLY FOR OTHER PERCENTAGES.

| Description Preserve | Localization Recall ($\varepsilon < 5/10/15m$)↑ | | |
|---|---|---|---|
| | k = 1 | k = 5 | k = 10 |
| 25% | 0.26/0.28/0.31 | 0.50/0.53/0.57 | 0.61/0.64/0.67 |
| 50% | 0.43/0.45/0.50 | 0.70/0.71/0.74 | 0.80/0.82/0.84 |
| 75% | 0.53/0.56/0.63 | 0.82/0.85/0.88 | 0.89/0.92/0.92 |
| 100% (CCCA alignment) | **0.56/0.59/0.65** | **0.85/0.86/0.89** | **0.91/0.92/0.94** |
| 100%(data alignment) | **0.57/0.60/0.66** | **0.86/0.87/0.89** | **0.92/0.93/0.94** |

TABLE VII

ROBUSTNESS STUDY WITH DIFFERENT VIEWPOINTS BASED ON THE TEST SET. WE EVALUATE THE IMPACT OF USING DIFFERENT NUMBERS OF IMAGES PER LOCATION ON THE MODEL'S ACCURACY. "1" INDICATES THAT FOR THE GIVEN LOCATIONS, ONLY A SINGLE IMAGE AND ITS CORRESPONDING TEXTUAL DESCRIPTION ARE USED FOR LOCALIZATION.

| Image Number | Localization Recall ($\varepsilon < 5/10/15m$)↑ | | |
|---|---|---|---|
| | k = 1 | k = 5 | k = 10 |
| 1 | 0.30/0.34/0.39 | 0.60/0.63/0.70 | 0.73/0.75/0.80 |
| 2 | 0.37/0.41/0.48 | 0.68/0.70/0.77 | 0.80/0.81/0.86 |
| 3 | 0.50/0.53/0.60 | 0.81/0.82/0.87 | 0.90/0.91/0.93 |
| 4 (CCCA alignment) | **0.56/0.59/0.65** | **0.85/0.86/0.89** | **0.91/0.92/0.94** |
| 4(data alignment) | **0.57/0.60/0.66** | **0.86/0.87/0.89** | **0.92/0.93/0.94** |

retained, the model's top-1 accuracy within a 5-meter range on the test set drops to 0.26. This accuracy improves to 0.43 when 50% of the content is retained, suggesting that more comprehensive descriptions enable more precise localization. Further analysis reveals that the top-1 accuracy difference between retaining 75% and 100% of text description is only 3% on the validation set and 2% on the test set. This indicates that 75% of the descriptions generally provide sufficient information for accurate localization, with additional text offering diminishing returns on accuracy.

*2) Robustness Study with Different Viewpoints:* To assess the effectiveness of multi-view matching, we evaluate the model using 1, 2, 3, and 4 images per location, as summarized in Table VII. When only a single image is used for matching, the top-1 accuracy on the test set reaches 0.30/0.34/0.39, indicating that our method maintains effectiveness and reliability even under single-view conditions. As the number of image perspectives increases, the model's accuracy improves consistently, which we attribute to each additional image expanding the field of view by approximately 90°, thereby capturing more contextual information. This experiment strongly supports the effectiveness of our multi-view inference strategy.

### F. Qualitative Evaluation

In addition to the quantitative results, we present qualitative results showcasing successful image localization from text descriptions and some failure cases in Fig. 4. Given a set of query text descriptions, we visualize the ground truth and the top 5 retrieved viewpoints. In text-based sub-viewpoint global place recognition, a retrieved sub-viewpoint is considered positive if it corresponds to the target location. Fig. 4 shows that Text4VPR achieves accurate localization through a multi-view matching approach.

In most cases, Text4VPR successfully retrieves the correct result in the top-1 localization. Additionally, within the top-5 retrieval results, Text4VPR frequently identifies multiple images that closely resemble the ground truth, as shown in Fig. 4(a) and Fig. 4(b). In instances of top-1 localization errors, illustrated in Fig. 4(c), the top-1 retrieved image is only six meters from the ground truth, while the ground truth appears in the top-2 position, still reflecting Text4VPR's high localization accuracy. Although some retrieved results are farther from the ground truth, these images exhibit a high degree of similarity to the textual description. Overall, our visualization results demonstrate that Text4VPR performs robustly in text-to-image localization tasks. Localization errors primarily arise from inherent ambiguities in semantic descriptions, which lead the system to match locations with highly similar images.

## V. CONCLUSION AND DISCUSSION

We present Text4VPR and build the first baseline for text-image place recognition with the Street360Loc, representing the first method and dataset specifically designed for text-to-image localization, thereby completing the cross-modal place recognition framework. Text4VPR utilizes multi-view images in the place recognition task by training on text-image pairs and performing inference through text group to image group retrieval after CCCA alignment, achieving a top-1 accuracy of 56% within a 5-meter range. As a pioneering approach, Text4VPR demonstrates the feasibility of text-to-image place recognition and reveals its significant potential. We hope our work will inspire further research into cross-modal place recognition, with a focus on advancing text-to-image localization.

**Potential limitations**: The multi-view approach for cross-modal place recognition faces limitations such as semantic discrepancies, scalability challenges, generalization difficulties, ambiguous textual descriptions, noisy visual data, and complex feature integration and evaluation metrics.

**Future work**: Future work in multi-view cross-modal place recognition should enhance data alignment, feature fusion strategies, and generalization methods. It must also address textual ambiguity, develop scalable architectures, and prioritize user-centric evaluation metrics, while promoting interdisciplinary collaboration for better contextual understanding.

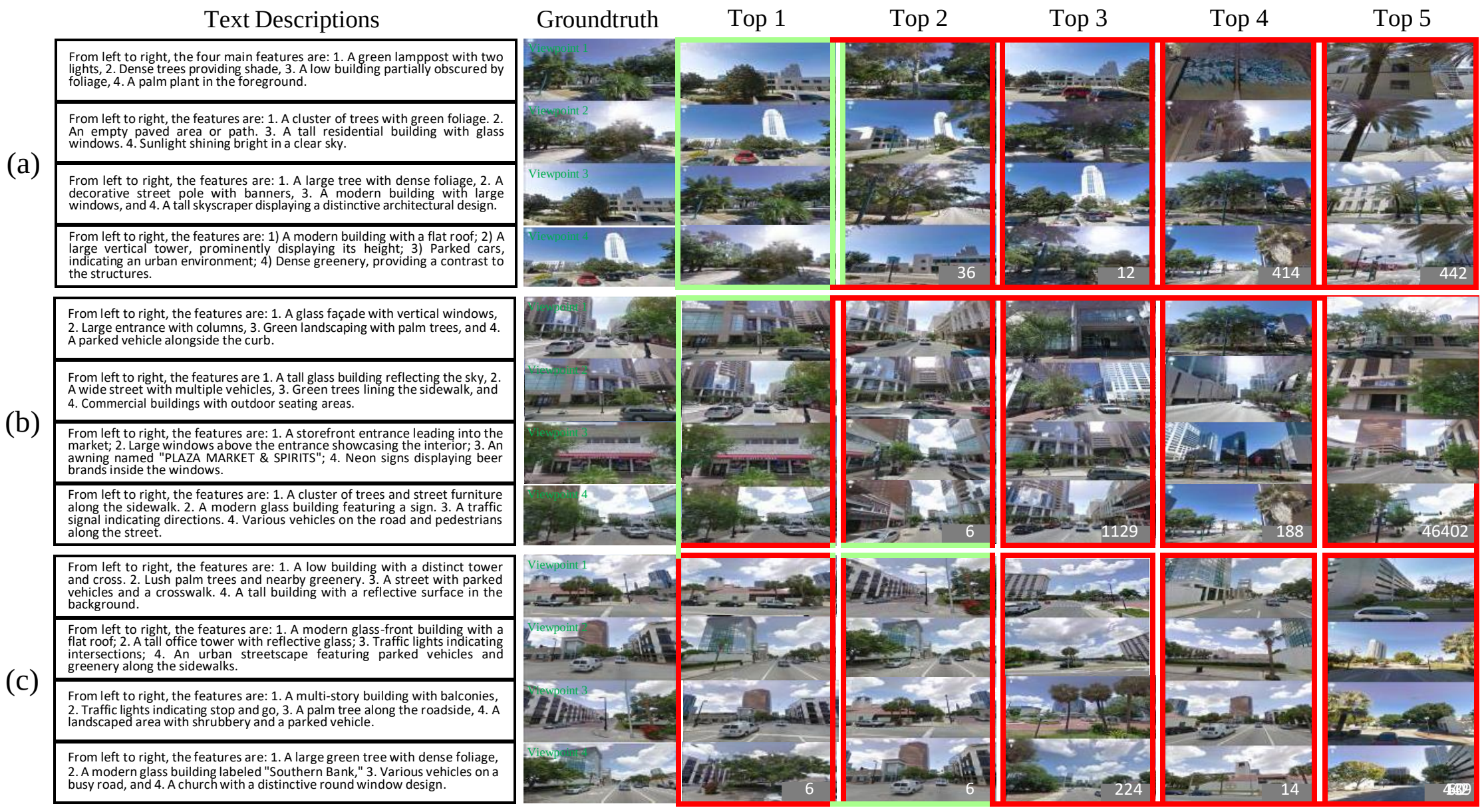

Fig. 4. Qualitative localization results on the Street360Loc dataset: In global place recognition, the numbers in the top-5 retrieval indicate the center distances ($m$) between the retrieved viewpoints and the ground truth (4 different views from 4 different directions). Green boxes represent positive viewpoints that contain the target location, while red boxes indicate negative views. Notably, the minimum distance between any two images in the dataset exceeds six meters. Note: the number in the lower right corner indicates the distance from the current localization to the ground-true.